\documentclass[a4paper,journal]{IEEEtran}
\usepackage{cite}
\usepackage{amsmath}
\usepackage{array}
\usepackage{amssymb}

\usepackage[pdftex]{graphicx}
\usepackage{caption}
\usepackage{subcaption}

\usepackage{float}
\usepackage{lineno,hyperref}
\usepackage{verbatim}

\usepackage{algorithmicx}
\usepackage[ruled]{algorithm}
\usepackage{algpseudocode}

\usepackage{comment}
\usepackage{lineno,hyperref}
\modulolinenumbers[5]
\usepackage{amsmath}
\usepackage{array}
\usepackage{amssymb}

\usepackage{algorithmicx}
\usepackage[ruled]{algorithm}
\usepackage{algpseudocode}
\newcommand{\myindent}[1]{
\newline\makebox[#1cm]{}
}
\usepackage{float}
\usepackage{lineno,hyperref}
\usepackage{verbatim}  
\usepackage{subcaption}

\begin{document}

\title{Progressive Graph Convolutional Networks\\for Semi-Supervised Node Classification}

\author{\IEEEauthorblockN{Negar Heidari and Alexandros Iosifidis}\\
\IEEEauthorblockA{Department of Electrical and Computer Engineering, Aarhus University, Denmark\\
Emails: \{negar.heidari, alexandros.iosifidis\}@ece.au.dk}}

\maketitle

\begin{abstract}
Graph convolutional networks have been successful in addressing graph-based tasks such as semi-supervised node classification. Existing methods use a network structure defined by the user based on experimentation with fixed number of layers and neurons per layer and employ a layer-wise propagation rule to obtain the node embeddings. Designing an automatic process to define a problem-dependant architecture for graph convolutional networks can greatly help to reduce the need for manual design of the structure of the model in the training process. 
In this paper, we propose a method to automatically build compact and task-specific graph convolutional networks. Experimental results on widely used publicly available datasets show that the proposed method outperforms related methods based on convolutional graph networks in terms of classification performance and network compactness.
\end{abstract}


\IEEEpeerreviewmaketitle


\section{Introduction}
\label{sec:introduction}
Convolutional Neural Networks (CNNs) \cite{lecun2015deep}, as an end-to-end deep learning paradigm, have been very successful in many machine learning and computer vision tasks \cite{krizhevsky2012imagenet, szegedy2015going, he2016deep}.
While CNNs have high ability to extract latent representations and local meaningful statistical patterns from data, they can only operate on Euclidean data structures such as audio, images and videos which have a form based on $1$D, $2$D and $3$D regular grids, respectively. Recently, there has been an increasing research interest in applying deep learning approaches on non-euclidean data structures, like graphs, which lack geometrical properties, but have high ability to model complex irregular data such as social networks \cite{hamilton2017inductive, kipf2016semi} citation networks and knowledge graphs \cite{hamaguchi2017knowledge}. 

Motivated by CNNs, the notion of convolution was generalized from grid data to graph structures, which correspond to locally connected data items, by aggregating the node's neighbors' features with its own features \cite{wu2019comprehensive}. Combining such convolutional operator with a data transformation process and hierarchical structures Graph Convolutional Networks (GCNs) are obtained, which can be trained in an end-to-end fashion to optimize an objective defined on individual graph nodes, or the graph as a whole. As an example, the GCN method proposed in \cite{kipf2016semi} is a semi-supervised node classification method consisting of two convolutional layers. Each graph convolutional layer learns nodes' features by applying aggregation rule on their corresponding first-order neighbors.
The graph structure can also be learned dynamically \cite{manessi2020dynamic}. 
GCNs, have been successful in various graph mining tasks like node classification, \cite{atwood2016diffusion, luo2020every}, graph classification \cite{zhang2018end}, link prediction \cite{zhang2018link}, and visual data analysis \cite{chen2020graph}. 

One of the main drawbacks of existing GCN-based methods is that they require an extensive hyper-parameter search process to determine a good topology of the network. This process is commonly based on extensive experimentation involving the training of multiple GCN topologies and monitoring their performance on a hold-out set. To avoid such a computationally demanding process, a common practice is to empirically define a network topology and use it for all examined problems (data sets). As an example, a two layer GCN model cannot gain sufficient global information by aggregating the information from just a two-hop neighborhood for each node. Meanwhile, in the reported results of \cite{kipf2016semi} it has been shown that adding more layers makes the training process harder and does not necessarily improve the classification performance. It was also shown that the topology of the GCN plays a crucial role in performance which is related to the underlying difficulty of the problem to be solved \cite{vignac2020choice}.

Problem-specific design of the neural networks' architecture contributes in improving the performance and the efficiency of a learning system. Recently, methods of finding an optimized network topology have been receiving much attention and many works were proposed to define compact network topologies by employing various learning strategies, such as compressing pre-trained networks, adding neurons progressively to the network, pruning the network's weights and applying weight quantization \cite{cheng2017survey, howard2017mobilenets,  tran2018improving, tran2019heterogeneous}. However, all these methods work with grid data in Euclidean spaces. In this regard, learning a compact topology for GCNs makes a step towards increasing the training efficiency and reducing the computational complexity and storage needed while achieving comparable performance with existing methods. 

In this paper, we propose a method to jointly define a problem-specific GCN topology and optimize its parameters, by progressively growing the network's structure both in width and depth. The contributions of our work are:

\begin{itemize}
	\item We propose a method to learn an optimized and problem-specific GCN topology progressively without user intervention. The resulting networks are compact in terms of number of parameters, while performing on par, or even better, compared to other recent GCN models. 
	\item We provide a convergence analysis for the proposed approach, showing that the progressively building GCN topology is guaranteed to converge to a (local) minimum.
	\item We conduct experiments on widely-used graph datasets and compare the proposed method with recently proposed GCN models to demonstrate both the efficiency and competitive performance of the proposed method. Our experiments include an analysis of the effect of the network's complexity with respect to the underlying complexity of the classification problem, highlighting the importance of network's topology optimization.
\end{itemize}

The rest of the paper is organized as follows: Section \ref{sec:background} provides a description of graph-based semi-supervised classification, along with the terminology used in this paper. Section \ref{sec:GCN} reviews the GCN method \cite{kipf2016semi} as the baseline method of our work. The proposed method is described in detail in section \ref{sec:PGCN}. The conducted experiments are described in section \ref{sec:experiments}, and conclusions are drawn in Section \ref{sec:conclusion}. 

\section{Semi-supervised graph-based classification}\label{sec:background}
Let $G = (\mathcal{V} ,\mathcal{E})$ be an undirected graph where $\mathcal{V}$ and $\mathcal{E}$ denote the set of nodes $\nu_{i}\in \mathcal{V}, \:i=1,\dots,N$ and the set of edges $(\nu_{i},\nu_{j})\in \mathcal{E}$, respectively. $\mathbf{A} \in R^{N\times N}$ denotes the adjacency matrix of $G$ encoding the node connections. The elements $\mathbf{A}_{ij}$ can be either binary values, indicating presence or absence of an edge between nodes $\nu_{i}$ and $\nu_{j}$, or real values encoding the similarity between $\nu_{i}$ and $\nu_{j}$, based on a similarity measure. Using $\mathbf{A}$, we define the degree matrix $\mathbf{D}$ which is a diagonal matrix with elements equal to $\mathbf{D}_{ii} = \sum _{j} \mathbf{A}_{ij}$. Each node of the graph $\nu_{i}$ is also equipped with a representation $\mathbf{x}_i \in \mathbb{R}^D$, which is used to form the feature vector matrix $\mathbf{X} \in R^{N\times D} $. When such a feature vector for each graph node is not readily available for the problem at hand (e.g. in the case of processing citation graphs), vector-based node representations are learned by using some node embedding method, like \cite{abu2017watch, grover2016node2vec}. 

Traditional graph-based semi-supervised node classification methods \cite{belkin2006manifold,weston2012deep,iosifidis2014regularized}, learn a mapping from the nodes' feature vectors to labels, which for a $C$-class classification problem are usually represented by $C$-dimensional vectors following the $1$-of-$C$ encoding scheme. This mapping exploits a graph-based regularization term and it commonly has the form:
\begin{equation}
\begin{aligned}
\mathcal{L} = \mathcal{L} _{L}\left(f(\mathbf{X}_L), \mathbf{T}_{L}\right) + \lambda f(\mathbf{X})^{T} \mathbf{L} f(\mathbf{X}),
\end{aligned}
\label{reg_GNN}
\end{equation}
where $f(\cdot)$ denotes the learnable function mapping the $D$-dimensional node representations to the $C$-dimensional class vectors, $\mathbf{T}_L$ is a matrix formed by the class label vectors of the labelled nodes which form the matrix $\mathbf{X}_L$, and $\mathbf{L} = \mathbf{D} - \mathbf{A}$ is the unnormalized graph Laplacian matrix. The first term in (\ref{reg_GNN}) is the classification loss of the trained model measured on the labelled graph nodes, and the second term corresponds to graph Laplacian-based regularization incorporating a smoothness prior to the optimization function $\mathcal{L}$. $\lambda > 0$ expresses the relative importance of the two terms. By following this approach the labels' information of the labelled graph nodes is propagated over the entire graph. 

\section{Graph Convolutional Networks}\label{GCN_subsec}
\label{sec:GCN}
GCNs are mainly categorized into spatial-based and spectral-based methods. 
The spatial-based methods update the features of each node by aggregating its spatially close neighbors' feature vectors. In these methods, the convolution operation is defined on the graph with a specified neighborhood size which propagates the information locally \cite{atwood2016diffusion, zhuang2018dual, monti2017geometric, hamilton2017inductive}. 

The spectral-based GCN methods follow a graph signal processing approach \cite{kipf2016semi}. Let us denote by $G_{\theta}$ a filter and by $\mathbf{X}$ a multi-dimensional signal defined over the $N$ nodes of the graph. The signal transformation using $G_{\theta}$ is given by: 
\begin{equation}
\begin{aligned}
G_{\theta} \star \mathbf{X} = \mathbf{U} G_{\theta} \mathbf{U}^{T} \mathbf{X},\label{Eq:GraphConvolution}
\end{aligned}
\end{equation}
where $\star$ denotes the convolution operator, $\mathbf{U}$ is the matrix of eigenvectors of the normalized graph Laplacian $\tilde{\mathbf{L}} = \mathbf{I}_N - \mathbf{D}^{-\frac{1}{2}} \mathbf{A} \mathbf{D}^{-\frac{1}{2}} = \mathbf{U} \boldmath{\Lambda} \mathbf{U}^T$ with $\boldsymbol{\Lambda}$ being a diagonal matrix having as elements the corresponding eigenvalues, and $\mathbf{U}^T \mathbf{X}$ being the graph Fourier transform of $\mathbf{X}$. Since computing the eigen-decomposition of $\tilde{\mathbf{L}}$ is computationally expensive, low-rank approximations using truncated Chebyshev polynomials have been proposed \cite{hammond2011wavelets}. The transformation in (\ref{Eq:GraphConvolution}) corresponds to the building block of a GCN. Followed by an element-wise activation it forms a GCN layer.

The multilayer GCN for semi-supervised node classification was proposed in \cite{kipf2016semi} by stacking multiple GCN layers. To achieve a fast and scalable operation a first-order approximation of the spectral graph convolution is proposed leading to:
\begin{equation}
\begin{aligned}
G_{\theta} \star \mathbf{X} \approx \tilde{\mathbf{D}}^{-\frac{1}{2}} \tilde{\mathbf{A}} \tilde{\mathbf{D}}^{-\frac{1}{2}} \mathbf{X} \mathbf{W},
\end{aligned}
\end{equation}
where $\tilde{\mathbf{A}} = \mathbf{I}_N + \mathbf{A}$ and $\tilde{\mathbf{D}}_{ii} = \sum_j \tilde{\mathbf{A}}_{ij}$. Let us denote by $\mathbf{H}^{(l)}$ the graph node representations at layer $l$ of the multi-layer GCN. The propagation rule for calculating the graph node representations at layer $l+1$ is given by:
\begin{equation}
\begin{aligned}
\mathbf{H}^{(l+1)} = \sigma\left(\tilde{\mathbf{D}}^{-\frac{1}{2}} \tilde{\mathbf{A}} \tilde{\mathbf{D}}^{-\frac{1}{2}} \mathbf{H}^{l} \mathbf{W}^{(l)}\right), 
\end{aligned}\label{eq:lap_reg}
\end{equation}
where $\mathbf{W}^{(l)}$ is the $l^{th}$ layer weight matrix and $\sigma(\cdot)$ denotes the activation function, such as $ ReLU(\cdot)$, or $softmax(\cdot)$ used for the output layer. For a two layer GCN model this leads to:
\begin{equation}
\begin{aligned}
\mathbf{Y} = softmax\Big(\hat{\mathbf{A}}  \: ReLU\left(\hat{\mathbf{A}} \mathbf{X} \mathbf{W}^{(1)} \right) \mathbf{W}^{(2)} \Big), 
\end{aligned}
\end{equation}
where $\hat{\mathbf{A}} = (\tilde{\mathbf{D}}^{-\frac{1}{2}} \tilde{\mathbf{A}} \tilde{\mathbf{D}}^{-\frac{1}{2}})$ and $\mathbf{Y} \in \mathbb{R}^{N\times C}$ denotes the predicted feature vectors for all the $N$ graph nodes in $C$ classes. The model parameters ($\mathbf{W}^{(0)},\mathbf{W}^{(1)}$) are finetuned by minimizing the cross entropy loss over the labeled nodes.

One of the drawbacks of all existing GCN methods is that they use a predefined network topology, which is selected either based on the user's experience, or empirically by testing multiple network topologies. In the next section, we describe a method for automatically determining a problem-specific compact GCN topology based on a data-driven approach.

\section{Progressive Graph Convolutional Network} \label{sec:PGCN}
PGCN follows a data-driven approach to learn a problem-dependant compact network topology, in terms of both depth and width, by progressively building the network's topology based on a process guided by its performance. That is, the learning process of GCN jointly determines the network's topology and estimates its parameters to optimize the cost function defined at the output of the network. 

The learning process starts with a single hidden layer formed by one block of $B$ neurons equipped with an activation function (e.g. $ReLU(\cdot)$) and an output layer with $ C $ neurons. At iteration $t=1$, the synaptic weight matrix $\mathbf{W}_{1}^{(1)} \in \mathbb{R}^{D\times B}$ connecting the input layer to the hidden-layer neurons is initialized randomly. The graph nodes' representations defined at the outputs of the hidden layer $\mathbf{H}_{1}^{(1)} \in \mathbb{R}^{N\times B}$, where the index indicates that the one block of $B$ hidden-layer neurons is used, are obtained by using graph convolution:
\begin{equation}
\mathbf{H}_{1}^{(1)}=ReLU\left(\hat{\mathbf{A}}\mathbf{X}\mathbf{W}_{1}^{(1)}\right).
\label{eq:block_hidden}
\end{equation}
By setting $\mathbf{H}^{(1)} = \mathbf{H}_{1}^{(1)}$, the network's output for all graph nodes $\mathbf{Y} \in \mathbb{R}^{N \times C}$ is calculated using a linear transformation as follows: 
\begin{equation}
    \mathbf{Y} = \mathbf{H}^{(1)} \mathbf{O}
    \label{eq:predict_firstL}
\end{equation}
where $\mathbf{O} \in \mathbb{R}^{B \times C}$ denotes the weight matrix connecting hidden layer to the output layer. The transformation matrix $\mathbf{O}$ can be calculated by minimizing the regression problem:
\begin{equation}
    \mathcal{J}_{1} = \frac{1}{2} Tr\left(\mathbf{O}^{T} \mathbf{O}\right) + \frac{\lambda_{1}}{2}\left \| \mathbf{H}_{L}^{(1)} \mathbf{O} - \mathbf{T}_{L} \right \|_{F}^{2}, 
\label{eq:ridge_regress}
\end{equation}
where $Tr(\cdot)$ denotes the trace operator, $\lambda_{1} > 0$ denotes the relative importance of model loss, $\mathbf{H}_{L}^{(1)}$ denotes the hidden layer representations of the labeled graph nodes and $\mathbf{T}_{L} \in \mathbb{R}^{N \times C}$ is a matrix formed by the labeled nodes' target vectors.

To exploit the information in both the labeled and unlabeled nodes' feature vectors in $\mathbf{H}^{(1)}$, we can replace the linear regression problem in (\ref{eq:ridge_regress}) by a semi-supervised regression problem exploiting the smoothness assumption of semi-supervised learning 
\cite{yan2006graph, iosifidis2014regularized} expressed by the term:
\begin{equation}
f(X)^{T}\mathbf{L} f(X)
= Tr\left(\mathbf{O}^T (\mathbf{H}^{(1)})^T \mathbf{\tilde{L}} \mathbf{H}^{(1)} \mathbf{O}\right).
\label{eq:lap_smoothing}
\end{equation}
Minimization of (\ref{eq:lap_smoothing}) with respect to $\mathbf{O}$ leads to node feature vectors at the output of the network which are similar for nodes connected in the graph. To incorporate this property in the optimization problem of (\ref{eq:ridge_regress}) the term in (\ref{eq:lap_smoothing}) is added as a regularizer. Thus, $\mathbf{O}$ is obtained by minimizing: 
\begin{equation}
\begin{aligned}
    \mathcal{J}_{2} = \frac{1}{2} Tr\left(\mathbf{O}^{T} \mathbf{O}\right) + \frac{\lambda_{1}}{2}\left \| \mathbf{H}_{L}^{(1)} \mathbf{O} - \mathbf{T}_{L} \right \|_{F}^{2} \\
    + \frac{\lambda_{2}}{2N^{2}} Tr\left(\mathbf{O}^T (\mathbf{H}^{(1)})^T \mathbf{\tilde{L}} \mathbf{H}^{(1)} \mathbf{O}) \right), 
\label{eq:optimization_func}
\end{aligned}
\end{equation}
where $N$ is the number of all (labeled and unlabeled) graph nodes and $\lambda_{2}$ denotes the relative importance of Laplacian regularization and is set to $ \frac{1}{N^{2}} $ which is the natural scale factor to estimate the Laplacian operator empirically \cite{belkin2006manifold}.

The optimal solution of (\ref{eq:optimization_func}) is obtained by setting ${\partial\mathcal{J}_{2}}/{\partial \mathbf{O}} = 0$, and is given by:
\begin{small}
\begin{equation}
\begin{split}
\mathbf{O} &= \left ( (\mathbf{H}^{(1)})^T \left(\mathbf{I}_N+\frac{\lambda _{2}}{\lambda _{1}N^{2}}\tilde{\mathbf{L}}\right)\mathbf{H}^{(1)}+\frac{1}{\lambda _{1}}\mathbf{I}_{D^{(1)}}
\right )^{-1} \\
&\cdot (\mathbf{H}_{L}^{(1)})^T\mathbf{T}_{L}
\end{split}
\label{eq:semi-reg}
\end{equation}
\end{small}
where $\mathbf{I}_{N} \in \mathbb{R}^{N \times N}$ and $\mathbf{I}_{D^{(1)}} \in \mathbb{R}^{{D^{(1)}} \times {D^{(1)}}}$ are identity matrices. ${D^{(1)}}$ denotes the feature dimension of the first hidden layer which is equal to $B$ in the first iteration. 

After the initialization of both the hidden layer and output layer weights, these are finetuned based on Backpropagation using the loss of the model on the labeled graph nodes. Finally, the model's performance (classification accuracy on the labeled nodes) $\mathcal{A}_{1}$ is recorded.

At iteration $t=2$, the network's topology grows by adding a second block of $B$ hidden layer neurons. We initialize the weights $\mathbf{W}_{1}^{(1)}$ connecting the input layer to the first block of hidden layer neurons with finetuned values obtained at iteration $t=1$, while the weight matrix of the newly added block $\mathbf{W}_{2}^{(1)}$ is initialized randomly. The hidden layer representations corresponding to the newly added neurons $\mathbf{H}_{2}^{(1)}$ are calculated as in (\ref{eq:block_hidden}) by replacing $\mathbf{W}_{1}^{(1)}$ with $\mathbf{W}_{2}^{(1)}$. Then, we combine $\mathbf{H}^{(1)} = [\mathbf{H}_{1}^{(1)}, \mathbf{H}_{2}^{(1)}]$ and the output weight matrix $\mathbf{O}$ is calculated by using (\ref{eq:semi-reg}). 
At this iteration, $D^{(1)} = B+B$ and $\mathbf{H}^{(1)}$, $\mathbf{O}$ are of size $N \times D^{(1)}$ and $D^{(1)} \times C$, respectively.

After finetuning all the parameters $\mathbf{W}_{1}^{(1)}$, $\mathbf{W}_{2}^{(1)}$, $\mathbf{O}$, and recording the model's performance, the network's progression is evaluated based on the rate of performance improvement given by:
\begin{equation}
\begin{aligned}
r=\frac{\mathcal{A}_{2}- \mathcal{A}_{1}}{\mathcal{A}_{1}},
\end{aligned}
\label{eq:node_prog}
\end{equation}
where $\mathcal{A}_{1}$ and $\mathcal{A}_{2}$ denote the model's performance before and after adding the second block of hidden layer neurons. 
If the addition of the second GCN block does not improve the model's performance, i.e. when $r < \epsilon$, the newly added block is removed, and all the model parameters are set to the finetuned values which were obtained at previous iteration. At this point, the progression in the current hidden layer terminates. 

After stopping the progressive learning in the first hidden layer, a new hidden layer is formed which takes as input the previous hidden layers' output. The block-based progression of the newly added hidden layer starts by using a single block of $B$ neurons and repeats in the same way as for the first hidden layer until model's performance converges.

\begin{figure*}[!t] 
\centering
\includegraphics[width=1.0\linewidth]{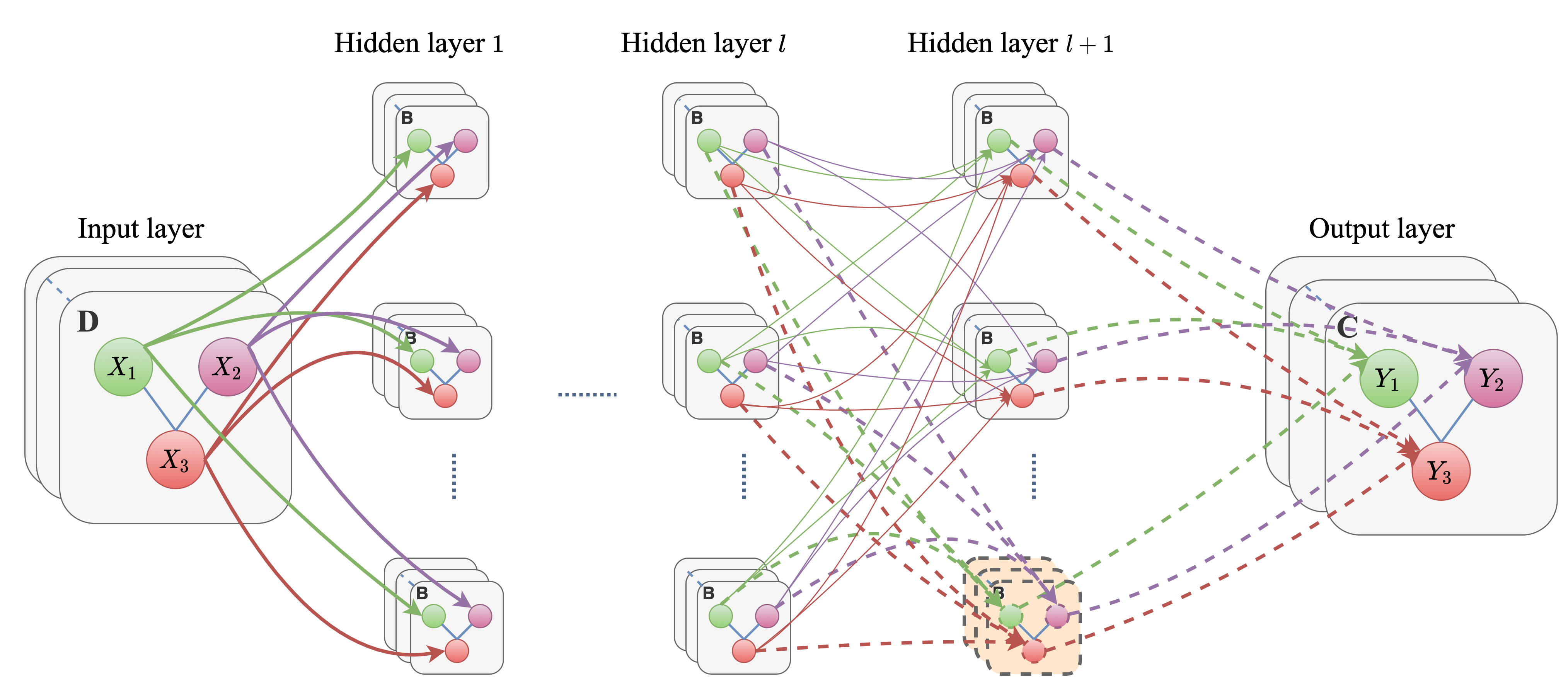}
\caption{PGCN model architecture, where $\mathbf{X}_{i}$ is the $i^{th}$ node feature vector and $\mathbf{Y}_{i}$ denotes the predicted label vector for $i^{th}$ node. $D$, $B$, $C$ indicate the node features' dimension, block size and number of classes respectively. Each layer consists of several blocks of neurons and each block contains $B$ neurons which apply graph convolution on the input graph data and transform the input features to $B$ dimensional space. 
When a new block of neurons is added to the ${(l+1)}^{th}$ layer, the weights shown by dashed lines which connect the $l^{th}$ layers' output to the newly added block are initialized randomly and the dashed line which connect the $l^{th}$ layers' hidden representation to the output layer are initialized using semi-supervised regression in (\ref{eq:semi-reg_general}). The solid lines indicate the finetuned weights of the $l^{th}$ layer. In order to evaluate the newly added block, all the model parameters are finetuned together.} 
\label{fig:model_diagram}
\end{figure*}

\begin{figure} 
\centering
\includegraphics[width=1.0\linewidth]{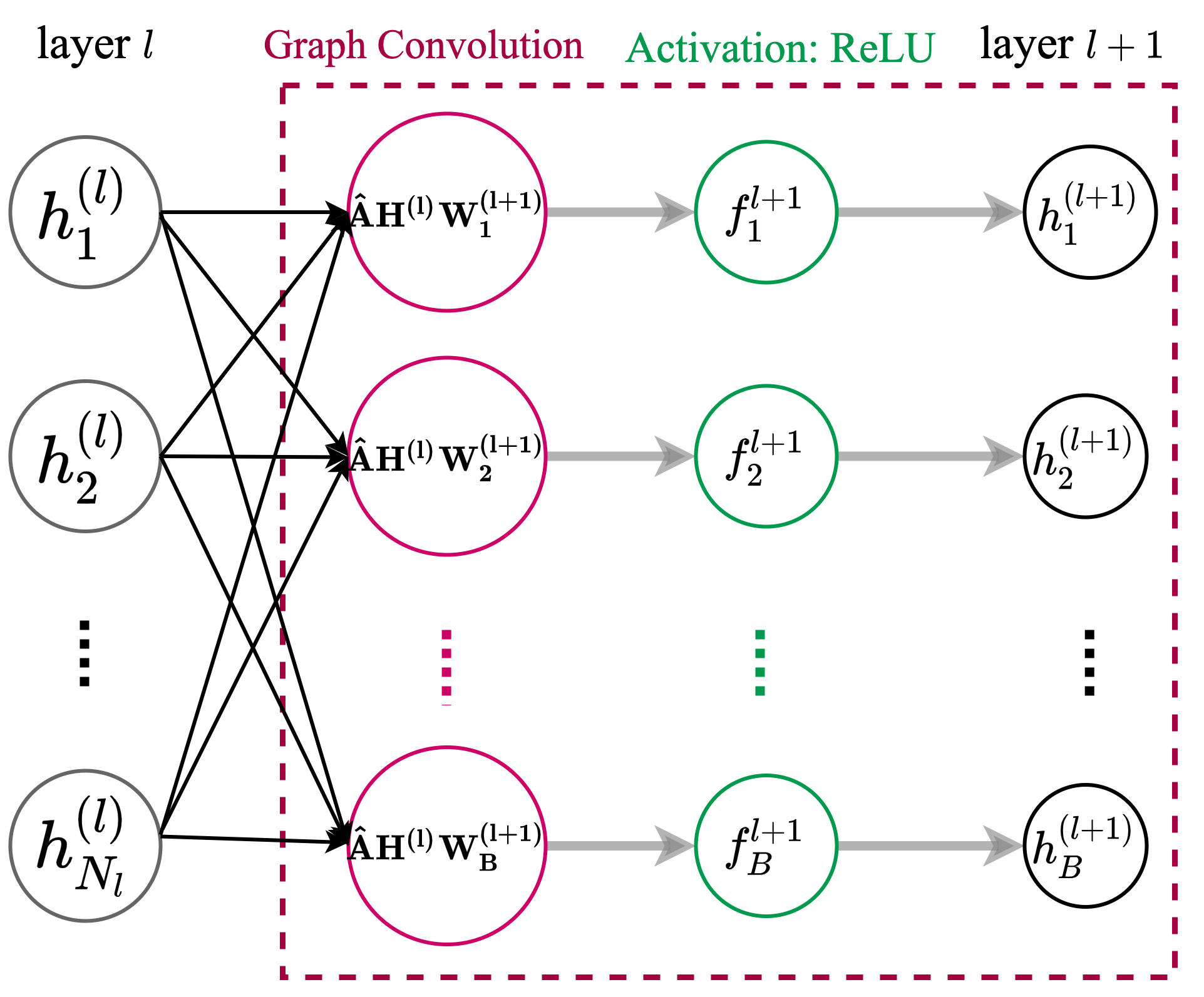}
\caption{The schematic of the graph convolution and activation operations which are applied on the hidden representation of each graph node in each block of the $l^{th}$ layer to produce its $B$ dimensional hidden representation in ${(l+1)}^{th}$ layer. Here, $\mathbf{\hat{A}}$ indicates the normalized adjacency matrix and $f_i$ is the activation function. ${h_i}^{(l)}$, ${h_i}^{(l+1)}$ indicate the output of each neuron in $l^{th}$ and ${(l+1)}^{th}$ layer, respectively and $\mathbf{H}^{(l)}$, $\mathbf{W}^{(l+1)}$ show the $l^{th}$ layer hidden representation, and the $(l+1)^{th}$ layer transformation matrix, respectively. 
} 
\label{fig:block_diagram}
\end{figure}

Let us assume that at iteration $t=k$, the network's topology comprises of $l$ layers (the input layer corresponds to $l=0$) and it is growing at the ${(l+1)}^{th}$ layer giving as outputs $\mathbf{H}^{(l+1)}$. Before adding the $b^{th}$ block formed by $B$ neurons, the weights of all the existing blocks in the network are set to the finetuned values obtained in the previous step. The newly added block in the ${(l+1)}^{th}$ hidden layer takes the output of the previous hidden layer $\mathbf{H}^{(l)}$ as input, and the graph convolutional operation for the $b^{th}$ block is given by: 
\begin{equation}
\mathbf{H}_{b}^{(l+1)}=ReLU\left(\mathbf{\hat{A}} \mathbf{H}^{(l)}\mathbf{W}_{b}^{(l+1)}\right)
\label{eq:hidden_rep_general}
\end{equation}
where $ \mathbf{H}_{b}^{(l+1)} $ and $ \mathbf{W}_{b}^{(l+1)}$ denote the hidden representation of the newly added block and the randomly initialized synaptic weights connecting the $l^{th}$ hidden layer to $b^{th}$ block of ${(l+1)}^{th}$ layer, respectively. 
Given $\mathbf{H}^{(l+1)} = [\mathbf{H}^{(l+1)}, \mathbf{H}_{b}^{(l+1)}]$ which denotes the hidden representations formed by using both the existing blocks and the newly added block in the ${(l+1)}^{th}$ hidden layer, the models' output $\mathbf{Y} \in \mathbb{R}^{N \times C}$ is calculated using a linear transformation as follows: 
\begin{equation}
    \mathbf{Y} = \mathbf{H}^{(l+1)} \mathbf{O}
    \label{eq:predict_general}
\end{equation}
where $\mathbf{O} \in \mathbb{R}^{D^{(l+1)} \times C}$ denotes the transformation matrix which is calculated based on semi-supervised linear regression:
\begin{small}
\begin{equation}
\begin{split}
\mathbf{O} &= \left ( (\mathbf{H}^{(l+1)})^T (\mathbf{I}_N +\frac{\lambda _{2}}{\lambda _{1}N^{2}}\tilde{\mathbf{L}})\mathbf{H}^{(l+1)} + 
\frac{1}{\lambda _{1}}\mathbf{I}_{D^{(l+1)}}\right)^{-1}\\
& \cdot (\mathbf{H}_{L}^{(l+1)})^T\mathbf{T}_{L}.
\label{eq:semi-reg_general}
\end{split}
\end{equation}
\end{small}

Similar to other GCN-based methods, the linear transformation for calculating the models' output in (\ref{eq:predict_general}) (and (\ref{eq:predict_firstL}), respectively) can also be followed by softmax activation (on a node basis) as follows:
\begin{equation}
    \mathbf{Y} = softmax\left(\mathbf{H}^{(l+1)} \mathbf{O}\right),
    \label{eq:predict_general_softmax}
\end{equation}
and instead of Mean Squared Error (MSE), the Cross-Entropy (CE) loss function can be employed for finetuning the model. 

After the initialization step, the synaptic weights of the existing blocks and all the weight parameters of the newly added block are finetuned with respect to the labeled data using (\ref{eq:ridge_regress}). 
The model diagram is shown in Fig. \ref{fig:model_diagram} which indicates the progressive learning process in ${(l+1)}^{th}$ layer where a new block is added and after initializing its weights (dashed lines), all the model parameters are finetuned together. Fig. \ref{fig:block_diagram} indicates the graph convolution process which is applied on graph node by a block of the $B$ neurons. This convolution process gets as input the hidden representation matrix of nodes at layer $l$, $\mathbf{H}^{(l)}$, and outputs the representation matrix $\mathbf{H}^{(l+1)}$ at layer $l+1$. 

To evaluate the network's progression, the model's performance is recorded and the rate of the improvement is given by:
\begin{equation}
\begin{aligned}
r_{b}=\frac{\mathcal{A}_{b}- \mathcal{A}_{b-1}}{\mathcal{A}_{b-1}}
\end{aligned}
\label{eq:block_conv}
\end{equation}
where $\mathcal{A}_{b-1}$ and $\mathcal{A}_{b}$ denote the classification accuracy before and after adding the $b^{th}$ block, respectively. 
If the addition of a new GCN block in step $ k $ does not improve the model's performance, i.e. when $r_{b} < \epsilon_{b} $, the progression of the $(l+1)^{th}$ layer terminates and all the network parameters are set to the finetuned values obtained in the previous step with $b-1$ blocks. 
After stopping the block progression for the $(l+1)^{th}$ layer, the algorithm evaluates whether the network's performance converged using the rate of the performance before and after adding the new hidden layer:
\begin{equation}
\begin{aligned}
r_{l+1}=\frac{\mathcal{A}_{l+1}- \mathcal{A}_{l}}{\mathcal{A}_{l}}.
\end{aligned}
\label{eq:layer_conv}
\end{equation}
When $r_{l} < \epsilon_{l}$, the last hidden layer $l+1$ is removed and the algorithm stops growing the network's topology. 
Here we should note that it is also possible to use other performance metrics, such as model loss, to evaluate the network progression process in (\ref{eq:block_conv}) and (\ref{eq:layer_conv}).

Algorithm \ref{alg:PGCN} summarizes the PGCN algorithm. In \ref{S:Appendix}, we show that the proposed approach of building GCN layers in a progressive manner converges.

\begin{algorithm}[!]
\caption{Progressive Graph Convolutional Network}
\label{alg:PGCN}
    \begin{algorithmic}[1]
    \Function{PGCN}{$\mathbf{X}$, $\mathbf{\tilde{A}}$, $\mathbf{\tilde{D}}$, $\mathbf{T}$, $\epsilon_{b}$, $\epsilon_{b}$, $l_{max}$ , $b_{max}$, $\alpha$, $\beta$}
    \State $\hat{\mathbf{A}}\leftarrow \mathbf{\tilde{D}}^{-\frac{1}{2}} \tilde{\mathbf{A}} \mathbf{\tilde{D}}^{-\frac{1}{2}}$ 
    \State $\tilde{L} \leftarrow \mathbf{I}_{N} - \mathbf{\tilde{D}}^{-\frac{1}{2}} \tilde{\mathbf{A}} \mathbf{\tilde{D}}^{-\frac{1}{2}}$
    \State $\mathbf{W}^{*} \leftarrow \:\{\}$, \:\: $\mathbf{O}^{*} \leftarrow \:\{\}$, \:\: $\mathbf{W}\leftarrow \:\{\}$
    \For{$l = 1$ to $l_{max}$} 
        \State $\mathbf{W}^{(l)}\leftarrow \:\{\}$
        \For{$b = 1$ to $b_{max}$} 
            \State Initialize $\mathbf{W}_{b}^{(l)}\sim \mathcal{U}(\alpha,\beta)$ 
            \State $\mathbf{\tilde{W}}^{(l)} = [\mathbf{W}^{(l)}, \mathbf{W}_{b}^{(l)}] $
            \State $\mathbf{H}_{b}^{(l)} \leftarrow \sigma(\mathbf{\hat{A}} \mathbf{H}^{(l)}\mathbf{W}_{b}^{(l)})$
            \State $\mathbf{H}^{(l)} = [\mathbf{H}^{(l)}, \mathbf{H}_{b}^{(l)}]$
            \State $\mathbf{O} \leftarrow \left( (\mathbf{H}^{(l)})^T (\mathbf{I}+\frac{\lambda _{2}}{\lambda _{1}N^{2}}\tilde{\mathbf{L}})\mathbf{H}^{(l)}+\frac{1}{\lambda_{1}}\mathbf{I} \right)^{-1} 
            \myindent{2} \cdot (\mathbf{H}_{L}^{(l)})^T \mathbf{T}_{L}$
            \State Finetune $\mathbf{\tilde{W}}^{(1:l)}$, $\mathbf{O}$
            \State Calculate $A_{b}^{new}$ 
            \State $r_{b} \leftarrow \frac{\mathcal{A}_{b}^{new}- \mathcal{A}_{b}^{old}}{\mathcal{A}_{b}^{old}}$
            \If{$r_{b}< \epsilon_{b}$} 
        	    \State break
    		\EndIf
    	   \State $\mathbf{W} \leftarrow \mathbf{\tilde{W}}$, \:\: $\mathcal{A}_{b}^{old} \leftarrow \mathcal{A}_{b}^{new}$
        \EndFor
        \State $r_{l} \leftarrow \frac{\mathcal{A}_{l}^{new}- \mathcal{A}_{l}^{old}}{\mathcal{A}_{l}^{old}}$
        \If{$r_{l} < \epsilon_{l}$}
            \State break
        \EndIf
    	\State $\mathbf{W}^* \leftarrow \mathbf{W}$, \:\: $\mathbf{O}^* \leftarrow \mathbf{O}$, \:\:  $\mathcal{A}_{l}^{old} \leftarrow \mathcal{A}_{l}^{new}$
    \EndFor 
    \State Return $\mathbf{W}^{*}$ , $\mathbf{O}^{*}$
    \EndFunction
    \Statex
    \end{algorithmic}
  \vspace{-0.4cm}
\end{algorithm}

\section{Experiments} \label{sec:experiments}
\subsection{Datasets}
We evaluated the proposed method for semi-supervised node classification task following transductive setting, on three widely-used benchmark datasets, Citeseer, Cora and Pubmed \cite{sen2008collective}, which are standard citation networks.

The citation networks represent published documents as nodes and the citation links between them as undirected edges. Each node in a citation network is represented by a sparse binary Bag-of-Words (BoW) feature vector extracted from articles' abstract and a class label which represents the articles' subject. The symmetric binary adjacency matrix $\mathbf{A}$ is built using the list of  undirected edges between the nodes and the task to be solved is the prediction of the articles' subject based on the BoW features and their citations to other articles. 

To perform a fair comparison, we follow the same experimental setup as in \cite{yang2016revisiting, kipf2016semi} for data configuration and preprocessing.  
The detailed datasets statistics are summerized in Table. \ref{table:datasets}.
For training the model, 20 labeled nodes per class are used for each citation network. In Table. \ref{table:datasets}, label rate denotes the number of training labeled nodes divided by the total number of nodes for each dataset. 
In all datasets, the validation set contains 500 randomly selected samples and the trained model is evaluated on 1000 test nodes.  
The labels of validation set are not used for training the model. Following the transductive learning setup, only the labeled nodes of the training set but all the feature vectors are used for training. The feature vectors are row-normalized. 

\begin{table}[!t]
	\centering
	\caption{Summary of datasets' statistics used in experiments.}\footnotesize
	\label{table:datasets}
	\begin{tabular}{llllll}
		\hline
		\cline{1-5}
		Dataset  & Citeseer & Cora & Pubmed \\
		\hline
		Type & Citation & Citation & Citation \\ 
		Nodes & 3327 & 2708& 19717 \\ 
		Edges & 4732 & 5429 & 44338 \\ 
		Classes & 6 & 7 & 3 \\ 
		Features & 3703 & 1433 & 500 \\ 
		Label rate  & 0.036 & 0.052 & 0.003 & \\ 
		\hline
	\end{tabular}
\end{table}

\subsection{Competing Methods}
We compared the proposed method with the baseline GCN \cite{kipf2016semi} and related methods N-GCN \cite{abu2018n}, GAT \cite{velivckovic2017graph}, GraphNAS \cite{gao2019graphnas} and Auto-GNN \cite{zhou2019auto}. 
N-GCN \cite{abu2018n} trains a network of GCNs over the neighboring nodes discovered at different distances in random walk. Each GCN module uses a different power of adjacancy matrix $\mathbf{A}$ like $\{ GCN(\hat{\mathbf{A}}^{0}), GCN(\hat{\mathbf{A}}^{1}), GCN(\hat{\mathbf{A}}^{2}), ..., GCN(\hat{\mathbf{A}}^{k})\}$, where $\hat{\mathbf{A}}^{k}$ indicates the statistics collected from the $k^{th}$ step of a random walk on the graph. It combines the information from different graph scales by using the weighted sum or the concatenation of all the GCNs' outputs into a final classification layer and finetunes the entire model for semi-supervised node classification. The graph attention network (GAT) \cite{velivckovic2017graph} introduces attention mechanism to GNNs. It assigns different weights to different neighboring nodes implicitly by employing self-attention through an end-to-end neural network architecture. The attention coefficients are shared across all graph edges, so there is no need to capture global graph structure. 

GraphNAS \cite{gao2019graphnas} and Auto-GNN \cite{zhou2019auto} are the most related works to ours which employ neural architecture search (NAS) mechanism to find the best graph neural network architecture automatically. GraphNAS utilizes a recurrent neural network (RNN) as a controller which generates the descriptions of different GNN architectures. The RNN is trained with reinforcement learning which gives feedback, like reward or penalty, to the controller in order to maximize the expected accuracy of the generated GNN architecture.  
Auto-GNN \cite{zhou2019auto} is also a NAS-based method which designed a novel parameter sharing method for GNN homogeneous architectures and uses an efficient RNN controller to capture the variations of data representation in GNN which are produced by small modifications made in GNN architecture during the search process. 

\subsection{Experimental settings}

We implemented our method in tensorflow \cite{abadi2016tensorflow} and trained it using Adam optimizer \cite{kingma2014adam} for $300$ epochs with learning rate of $0.01$. The network weight parameters are initialized randomly using uniform distribution. To handle the effect of randomness on network performance, we ran it $20$ times on each dataset. For each dataset, the set of hyper-parameters which leads to best validation performance is selected and the corresponding performance on test set and architectural information are reported. To avoid overfitting, dropout \cite{srivastava2014dropout} and regularization techniques are employed. The $L_2$ regularization factor is set to $0.0005$.
The dropout rate is selected from $\left \{ 0.1, 0.3, 0.5 \right \}$ and we apply dropout on the output of hidden layers, not on input features. The regularizer $\lambda_{1}$ is selected from $\left \{ 10^{-1}, 1, 10^{1} \right \}$ and the size of the block $B$ is selected from $\left \{1,5,10,15,20 \right \}$. The maximum topology of the network is limited to $10$ layers with $100$ neurons per layer and the threshold values are set to $\epsilon_l = \epsilon_b = 0.0001$. 

The baseline and competing methods also optimized the hyper-parameters on the same data splits. 
The test performance of these methods are reported directly from their papers.
GCN method optimized the hyper-parameters on Cora dataset and used the same set of hyper-parameters for Citeseer and Pubmed datasets too. The two layer GCN is trained for 200 epochs using Adam optimizer with a learning rate of $0.01$ and early stoping of step size $10$.
GCN with dropout rate of $0.5$, hidden layer of $16$ neurons and ${L}_{2}$ regularization of $0.0005$ is applied on citation datasets. 

N-GCN uses Adam optimizer with learning rate of $0.01$ for $600$ epochs and $ L_{2} $ regularization factor is set to $0.001$. This method uses a fixed architecture for all datasets. It has $4$ blocks of GCN which use different powers of $\hat{A}$: $\hat{A}^{0}$, $\hat{A}^{1}$, $\hat{A}^{2}$, $\hat{A}^{3}$. Each GCN, as a two layer network with $10$ hidden neurons, is replicated three times and the weighted sum of the outputs is introduced to the final classification layer. Similar to our experimental settings, N-GCN is run using 20 different random initializations and the test performance of the model with the highest accuracy on the validation set is reported. 

In GAT method, the hyper-parameters are optimized on Cora dataset and then reused for Citeseer dataset. The $ L_{2} $ regularization factor and the dropout rate are set to $0.0005$, $0.6$ respectively for Citeseer and Cora datasets. For pubmed dataset the $ L_{2} $ regularization factor is set to $0.001$. The dropout is applied on both input feature vectors and hidden layers output. 
GAT uses a two layer network architecture. The first layer consists of $8$ attention heads, each computing $8$ features, followed by ELU \cite{clevert2015fast} activation function and the second layer is a single attention head computing $C$ features followed by softmax activation function. The reported results for GAT method is the average of the classification accuracy on the test set after 100 runs. 

The NAS-based methods, GraphNAS and Auto-GNN, train a one-layer LSTM with 100 neurons as a controller using Adam optimizer with a learning rate of $0.0035$. The models which are sampled by the controller are trained for 200 epochs with the same hyper-parameter setting used in N-GCN. These methods explore only 2-layer GCN models in their predefined search space which leads to a constrained representation learning capacity for model. 
Both methods totally explore $1000$ different architectures to find the optimal graph neural architecture which achieves the best classification accuracy on the validation set. The reported results for GraphNAS method is the average performance of the top $5$ architectures on test set and the Auto-GNN method randomly initializes the best found architecture $5$ times and reports the average classification accuracy on test set. 
The Auto-GNN method is able to train the models with parameter sharing by transferring the finetuned weights, to a newly sampled architecture while the GraphNAS doesn't employ parameter sharing mechanism and it has a high computational and time complexity for training all 1000 different models from scratch. The results of Auto-GNN method both with and without using parameter sharing mechanism are reported in Table. \ref{table:Acc}. 

\subsection{Results}
Table. \ref{table:Acc} shows the performance in terms of classification accuracy for all the methods on the $3$ datasets. The best performance is shown in bold fonts for each dataset. 
We evaluate our method with both Cross Entropy (CE) and Mean Squared Error (MSE) loss functions. 
The obtained results in Table. (\ref{table:Acc}) indicate that the proposed method has outperformed the baseline GCN method and other competing methods on Citeseer and Pubmed datasets and it has competitive performance with competing methods on Cora dataset. 
In order to compare the efficiency of our method with competing methods, especially the NAS-based methods, the model sizes, i.e., the number of model parameters, of all trained models are reported in Table. \ref{table:params}. Table. \ref{table:arch} also shows the model architectures which are learned by our proposed method and the fixed model architectures used by the baseline method GCN. 
The results indicate that the network topologies which are learned by the proposed method on all datasets are much more compact compared to the fixed network topologies used by baseline methods, while the classification performance of our compact model is similar or better than others. 
Although the NAS-based methods achieve competitive performance, exploring $1000$ different architectures even by using a parameter sharing mechanism leads to an extremely high computation and time cost. To make their application tractable, their practical implementations restrict the search space by fixing the number of hidden layers to $2$. Thus, there is no guarantee that they can reveal an optimal topology of the network for achieving a good classification performance on large and complex graph datasets. Contrary to that, the proposed method can efficiently determine the network's topology by determining both the number of layers and neurons per layer.

\begin{table*}[!t]
	\centering
	\caption{Node classification performance in terms of accuracy.}
	\label{table:Acc}
	\begin{tabular}{llll}
		\hline \cline{1-4}
		Method  & Citeseer & Cora & Pubmed \\
		\hline
		GCN \cite{kipf2016semi} & $70.3$ & $81.5$ &  $79$ \\  
		N-GCN \cite{abu2018n} & $72.2$ & $83$ &  $79.5$ \\ 
		GAT \cite{velivckovic2017graph} &  $72.5\pm 0.7$ & $83.0\pm 0.7$ & $79.0\pm 0.3$ \\ 
		GraphNAS \cite{gao2019graphnas} & $73.1\pm 0.9$ & $\textbf{84.2} \pm \textbf{1.0}$ & $79.6\pm 0.4$ \\ 
		Auto-GNN \cite{zhou2019auto} & $73.8\pm 0.7$ & $83.6\pm 0.3$ & $79.7\pm 0.4$ \\ \hline
		PGCN (CE) & $\textbf{74.3}$ & $83.1$ & $\textbf{80.2}$ \\ 
		PGCN (MSE) & $72.7$ & $82.7$ & $80$ \\ \hline 
	\end{tabular}
\end{table*}

\begin{table}[!t]
	\centering
	\caption{Number of learnable parameters.}
	\label{table:params} 
	\begin{tabular}{llllllr}
		\hline \cline{1-4}
		Method  & Citeseer & Cora & Pubmed \\
		\hline
		GCN & 59.4k &  23.1k & 8.1k \\
		N-GCN & 445.4k & 173k & 60.5k \\ 
		GAT & 237.5k & 92.3k & 32.3k \\ 
		GraphNAS \cite{gao2019graphnas} & 230k & 188k & 30k \\ 
		Auto-GNN \cite{zhou2019auto} & 710k & 50k & 70k \\ \hline
		PGCN (CE) & \textbf{56k} & \textbf{14.7k} & \textbf{5.04k} & \\ 
		PGCN (MSE) & 74.3k & 22.1k & 7.5k & \\ 
		\hline
	\end{tabular}
\end{table}

\begin{table*}[!t]
	\begin{center}
	\caption{Network architectures comparison.}
	\label{table:arch} 
	\begin{tabular}{l|l|l|l}
		\hline \cline{1-3}
		Method  & GCN & PGCN (CE) & PGCN (MSE)\\
		\hline
		Citeseer & [D, h1: 16, 6] & [D, h1: 15, h2: 20, 6] & [D, h1: 20, 6]\\ 
		Cora &  [D, h1: 16, 7] & [D, h1: 10, h2: 20, 7] & [D, h1: 15, h2: 25, 7]\\
		Pubmed & [D, h1: 16, 3] & [D, h1: 10, 3] & [D, h1: 15, 3]\\
		\hline
	\end{tabular}
	\end{center}
	
	\begin{center}
	\begin{tabular}{l}
    	$D$ denotes the dimensionality of the input data\\
    	hX denotes the $X^{th}$ hidden layer.
	\end{tabular}
	\end{center}
	
\end{table*}

\begin{figure}
\centering
\begin{subfigure}[b]{0.35\textwidth}
\includegraphics[width=\textwidth]{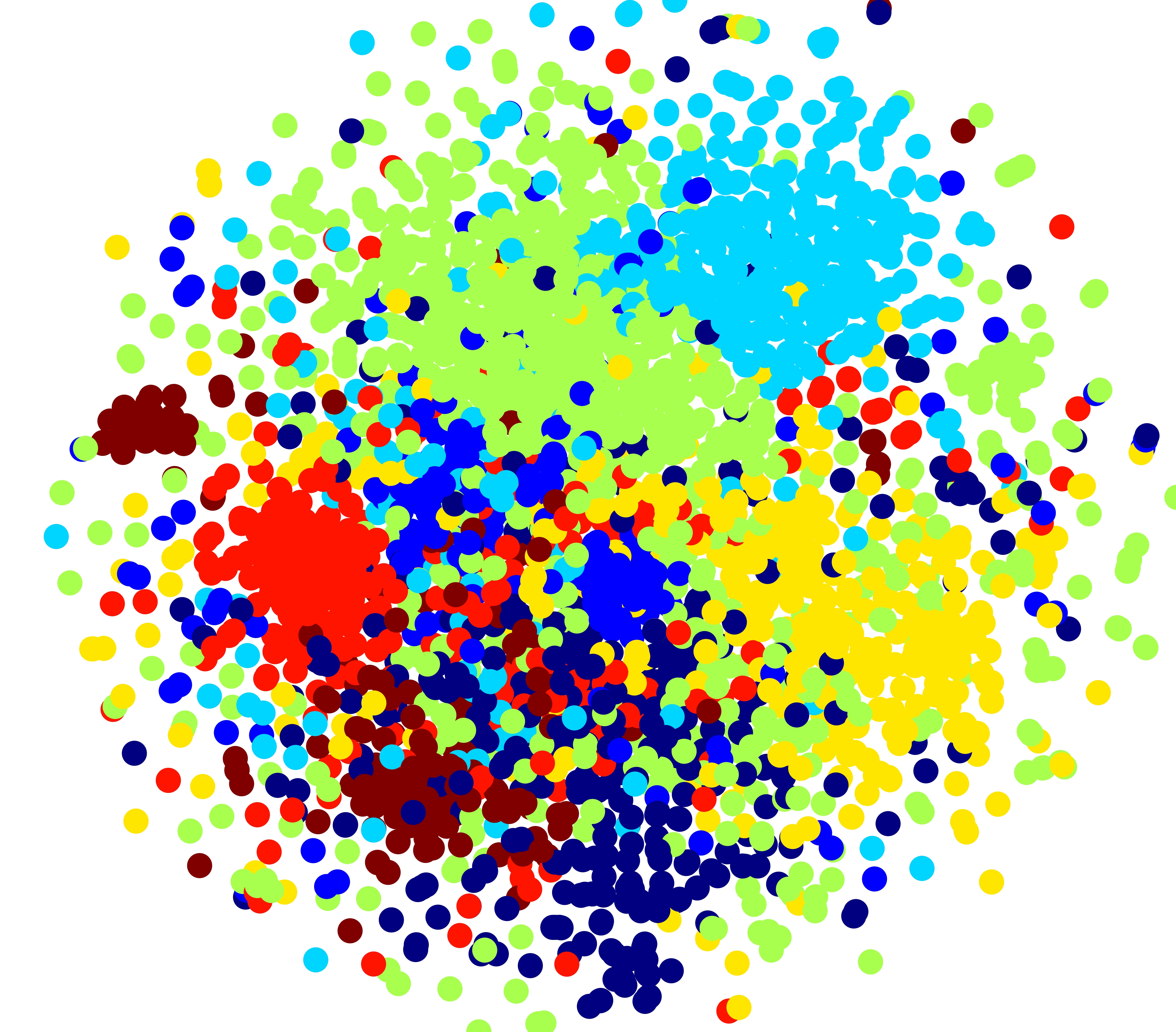}
\caption{Input node features} 
\end{subfigure}
\begin{subfigure}[b]{0.35\textwidth}
\includegraphics[width=\textwidth]{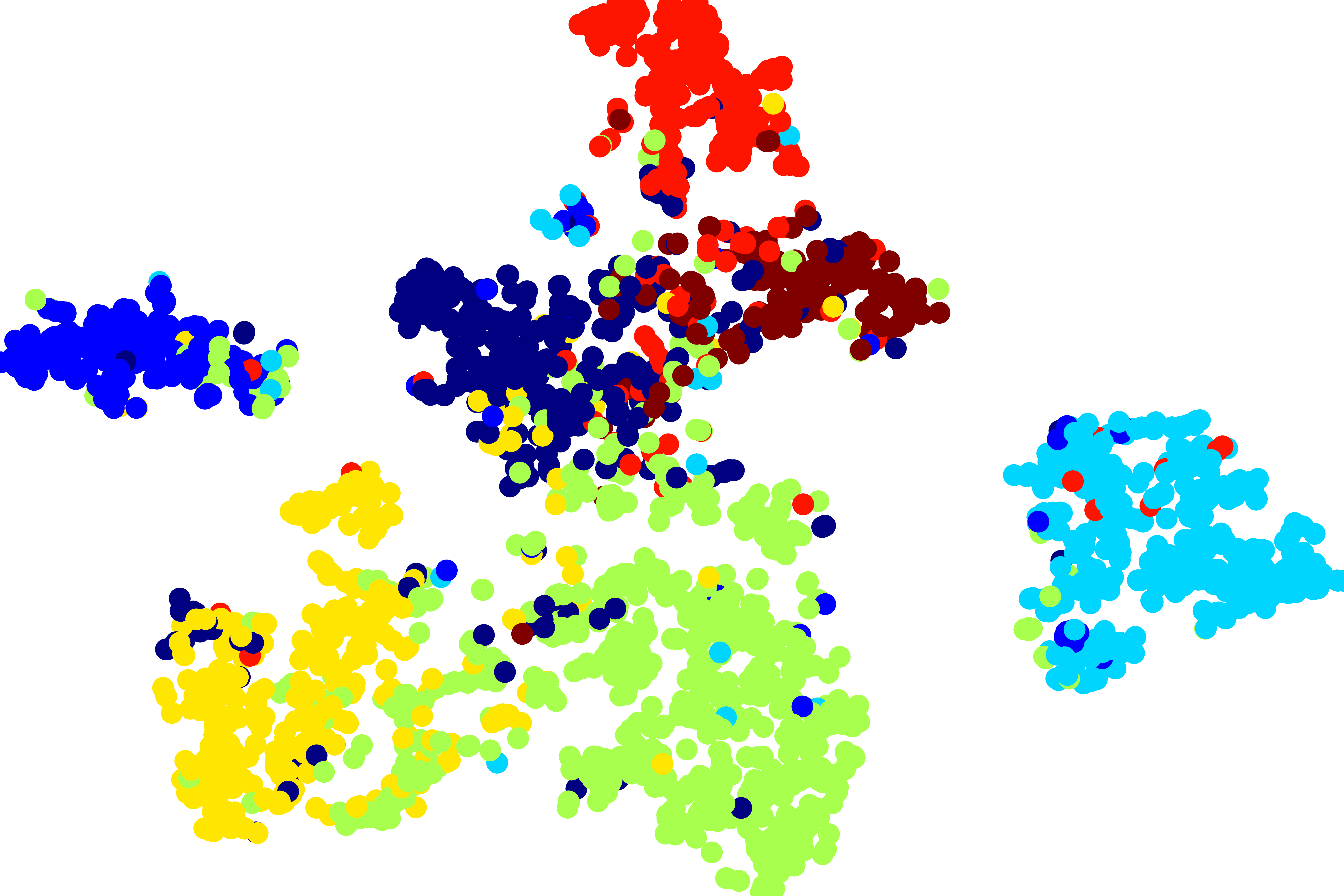}
\caption{Pre-softmax node features} 
\end{subfigure}
\caption{Demonstraion of Cora dataset node features visualized by t-SNE.} 
\label{fig:cora}
\end{figure}

Fig. \ref{fig:cora} illustrates the t-SNE visualization of learned feature vectors of the Cora dataset from last layer of network before applying the softmax activation.

\subsection{Analysis on Dataset Statistics}
The results of the previous section indicate that all the competing methods perform on par with the baseline GCN which has a simple network architecture. This can be explained by the benchmark dataset statistics. The dataset complexity is defined by the number of labeled nodes $N_{L}$ and the dimentionality of nodes' feature vectors $D$, and the ratio $N_{L}/D$ is extremely small for all the benchmark datasets of Table. {\ref{table:datasets}}. 
It has been recently shown that even the heavily regularized linear methods can obtain high performance on classification problems on datasets with low complexity \cite{wu2019simplifying, nt2019revisiting, klicpera2018predict}. Therefore, all the GCN-based methods, with simple or sophisticated network structures, can lead to comparable performance on these widely used benchmark datasets. In  \cite{vignac2020choice}, it has been experimentally shown that the performance of the GCN-based methods heavily depends on the underlying difficulty of the problem and non-linear models with more complex structure perform significantly better on datasets with higher $N_{L}/D$ ratio. That is, it is expected that the difference in performance of various methods will increase when the underlying semi-supervised classification problem becomes more complex.

Here we highlight the importance of optimizing the network structure based on the problem's complexity. We compare the performance of the proposed method with the baseline method GCN \cite{kipf2016semi}, by tuning the ratio $N_{L}/D$ using different input data dimensionalities when $50\%$ of nodes are labeled. We use the same data splits for both methods and follow the same approach as in \cite{vignac2020choice} to control the ratio $N_{L}/D$ by mapping the input data representations to a subspace through random projections. Specifically, we use a random sketching matrix $\mathbf{M}_{r} \in \mathbb{R}^{D \times D'}$ with $D' < D$, which is drawn from a Normal distribution, to obtain new data representations $\mathbf{X}^{\prime} \in \mathbb{R}^{D'}$ as follows: 
\begin{equation}
    \begin{aligned}
        \mathbf{X}^{\prime} = \frac{1}{\sqrt {D^{\prime}}} \mathbf{X} \mathbf{M}_{r}.
    \end{aligned}
\end{equation}
To avoid bias of the performance values obtained for different values of $D'$, we first randomly sample a square matrix $\mathbf{M}^{\prime} \in \mathbb{R}^{D \times D}$ and subsequently we use its first $r = D'$ columns to map the input data from $\mathbb{R}^D$ to its subspace $\mathbb{R}^{D'}$. Such an approach guarantees that when a subspace of a higher dimensionality is used, it corresponds to an augmented version of the initial (lower-dimensional) subspace. We applied three experiments for each choice of $D^{\prime}$ on each dataset and we report the performance on the test data corresponding to the best validation performance. 

\begin{figure*}[h]
\centering
    \begin{subfigure}[b]{\textwidth}
        \includegraphics[width=0.33\linewidth]{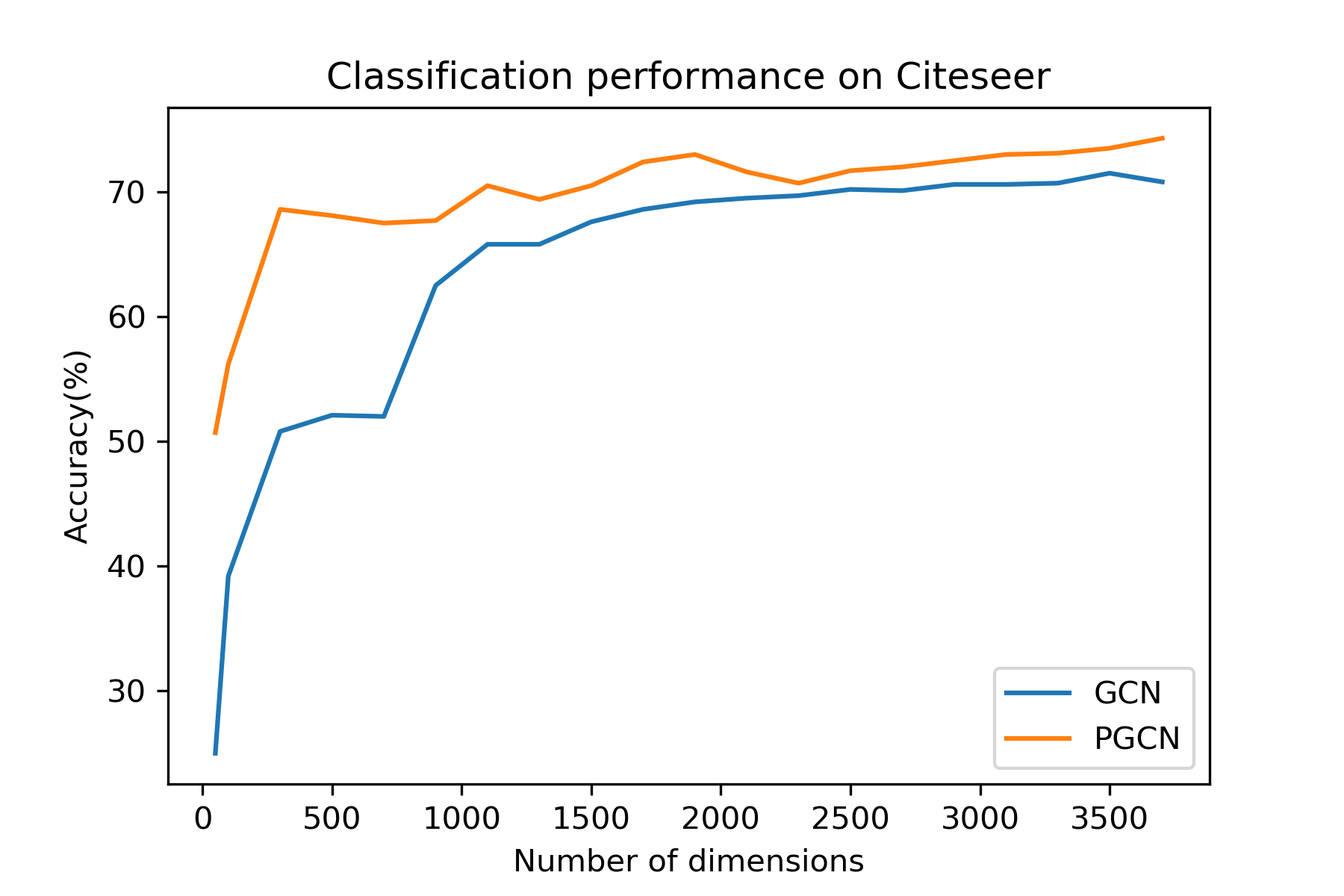}\hfil
        \includegraphics[width=0.33\linewidth]{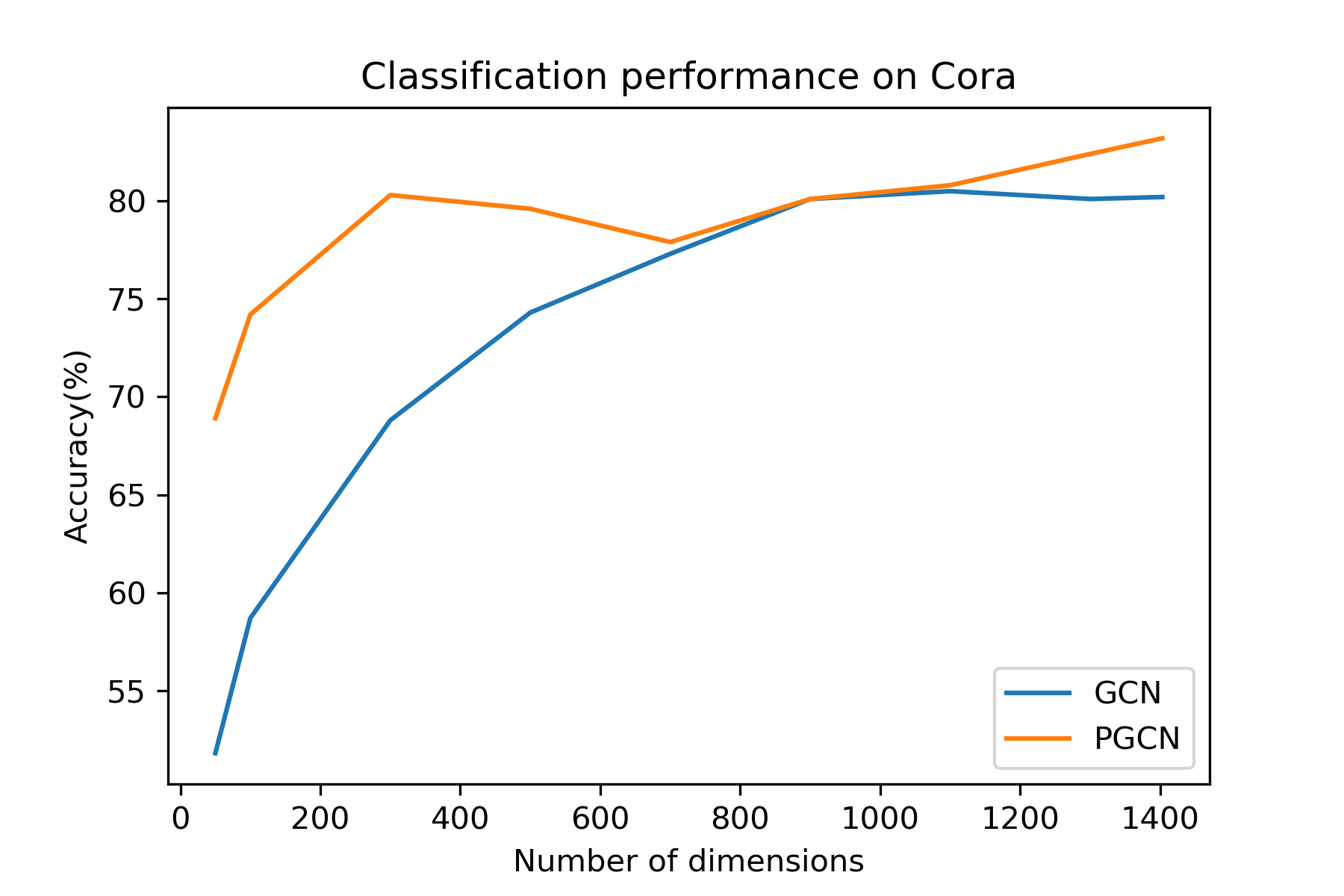}\hfil 
        \includegraphics[width=0.33\linewidth]{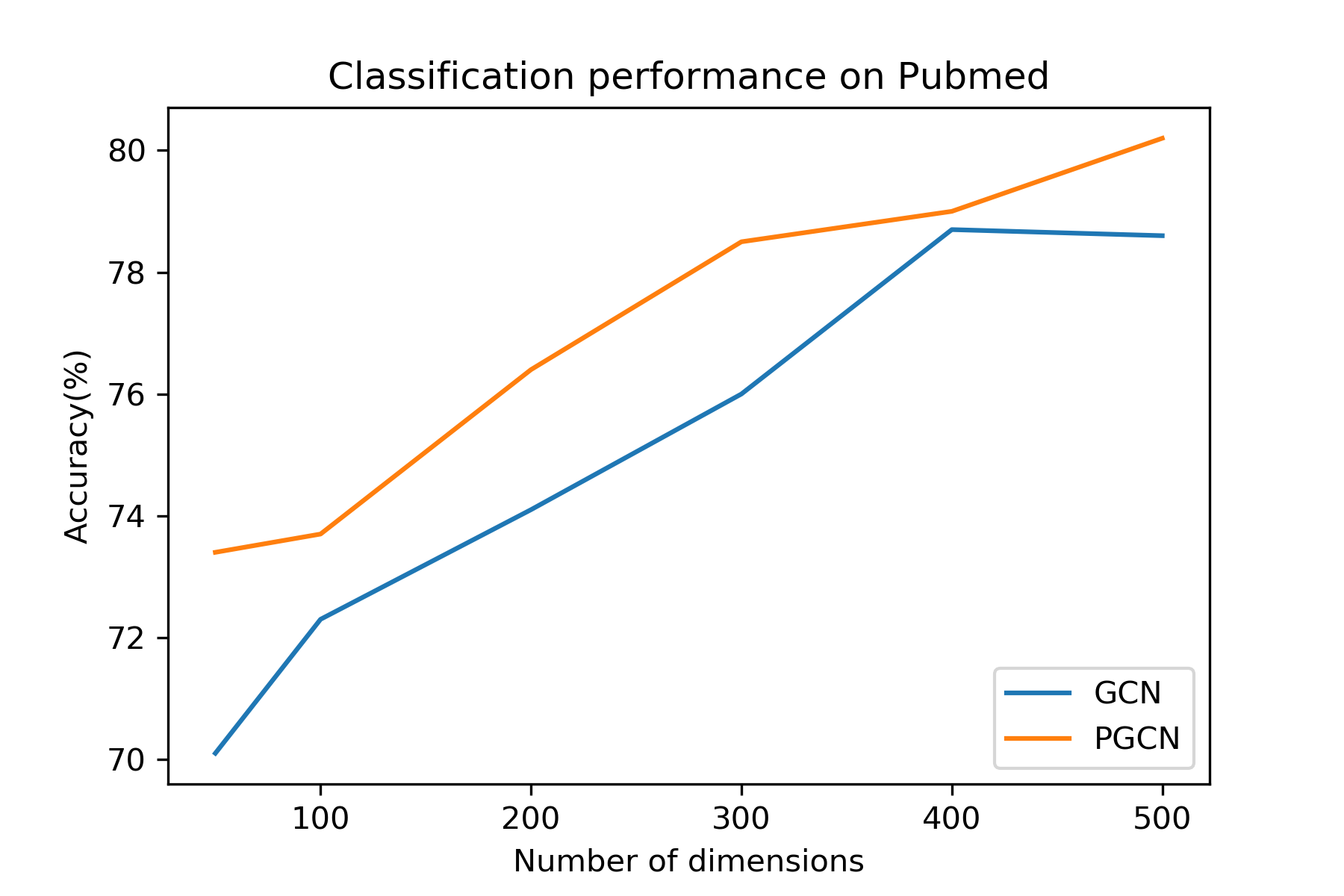}\hfil
        \caption{Classification performance comparison between GCN and PGCN in different dimensions}
    \end{subfigure}
    \begin{subfigure}[b]{\textwidth}
        \includegraphics[width=0.33\linewidth]{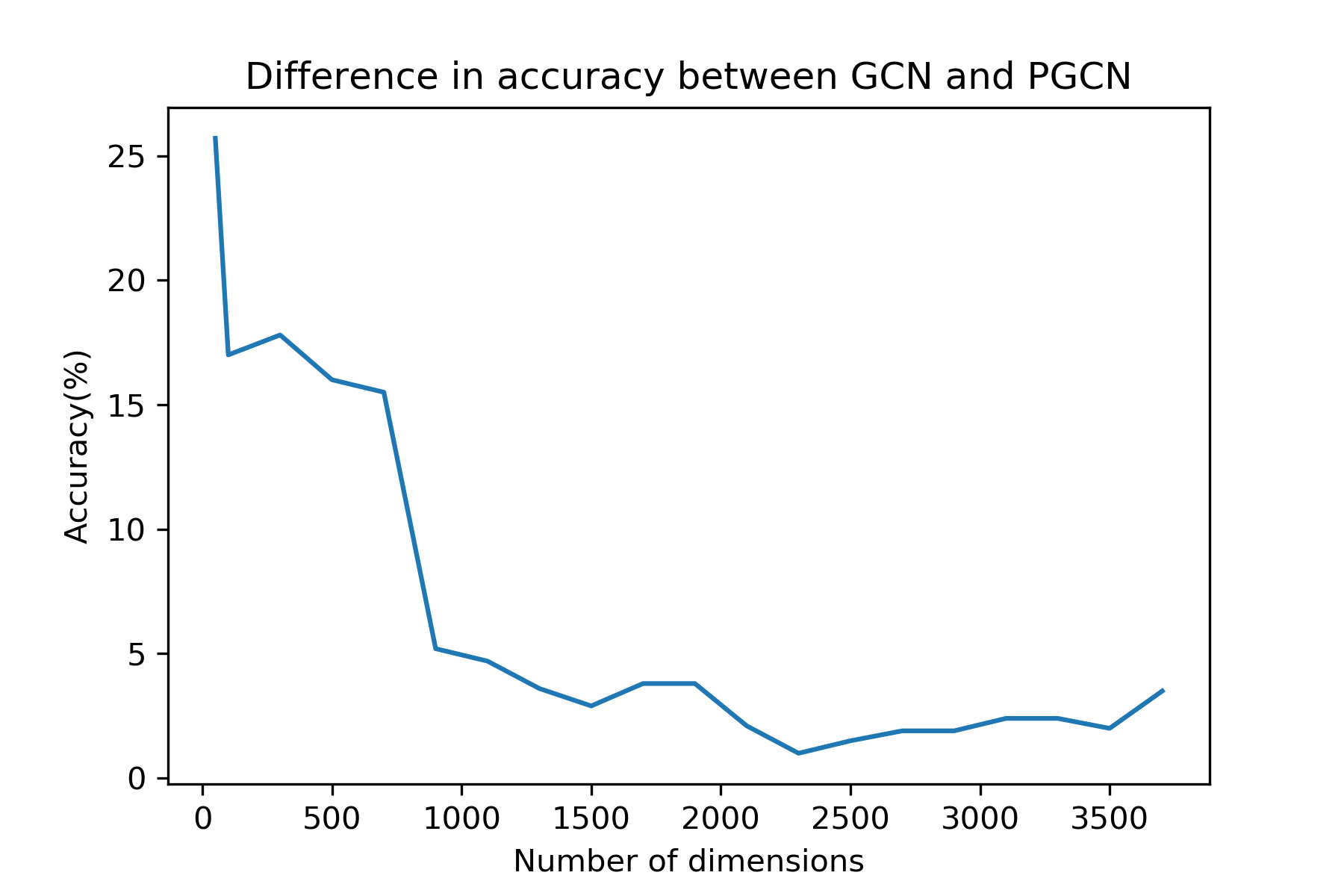}\hfil
        \includegraphics[width=0.33\linewidth]{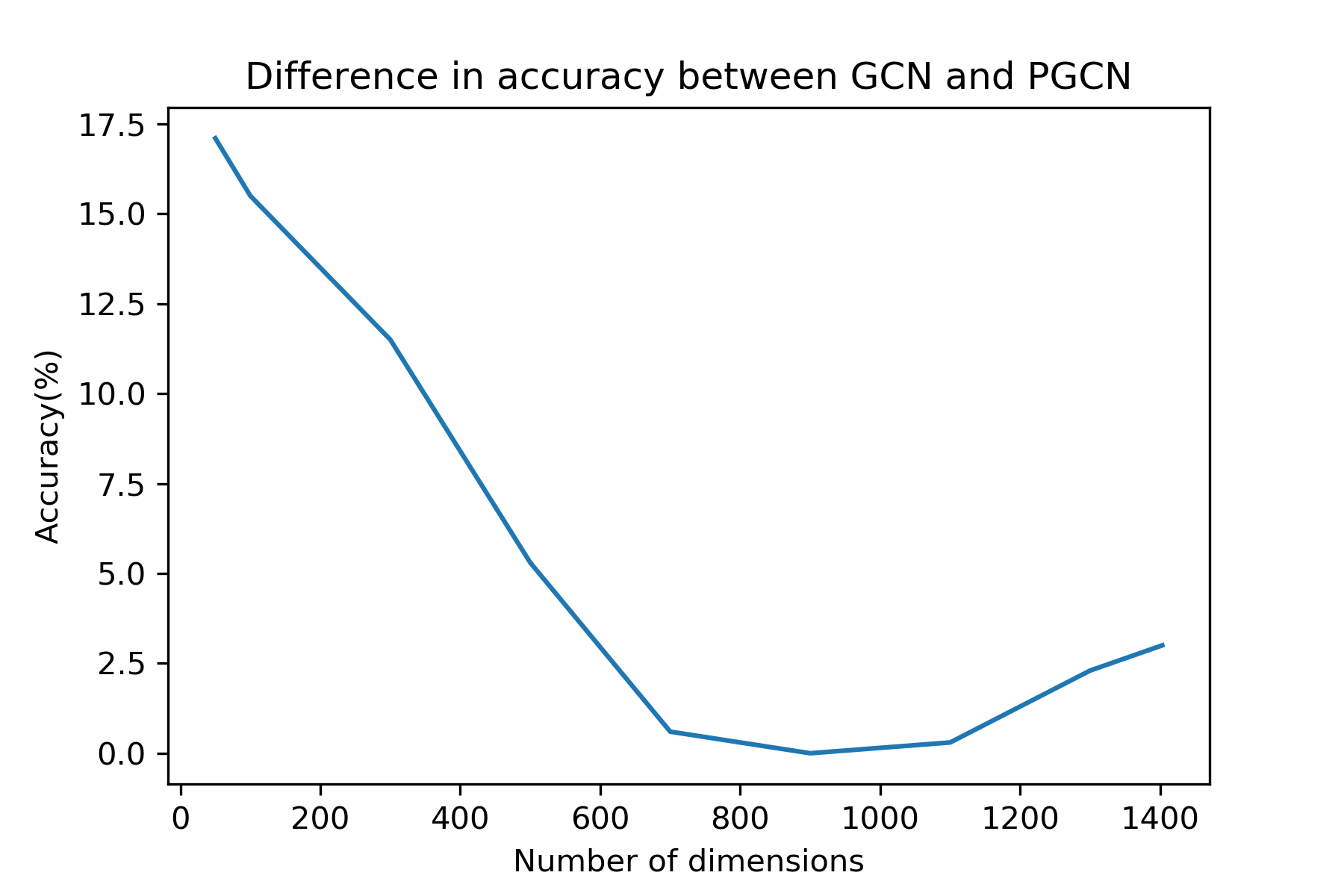}\hfil 
        \includegraphics[width=0.33\linewidth]{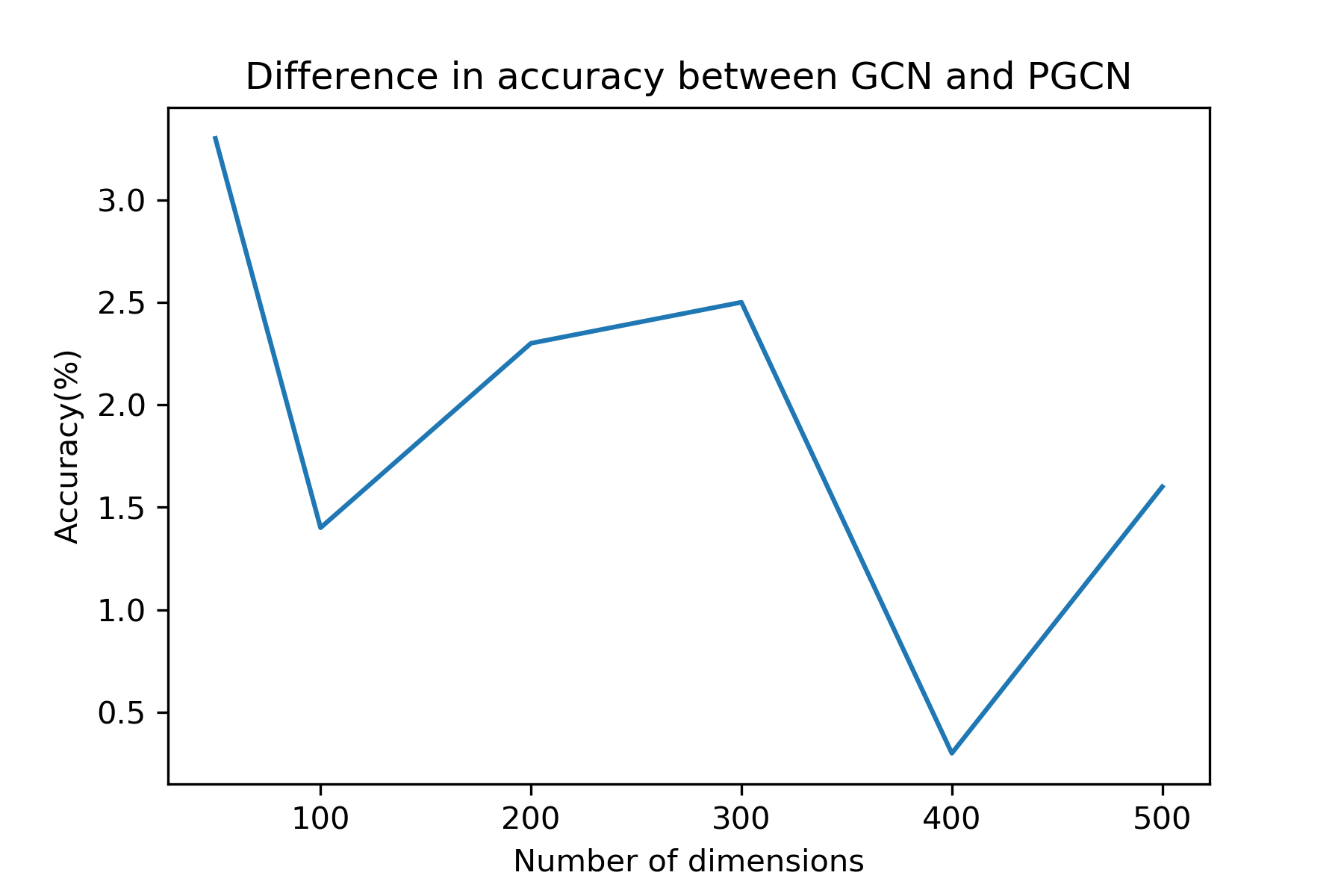}\hfil 
        \caption{Difference in classification accuracy between GCN and PGCN in different dimensions}
    \end{subfigure}
    \begin{subfigure}[b]{\textwidth}
        \includegraphics[width=0.33\linewidth]{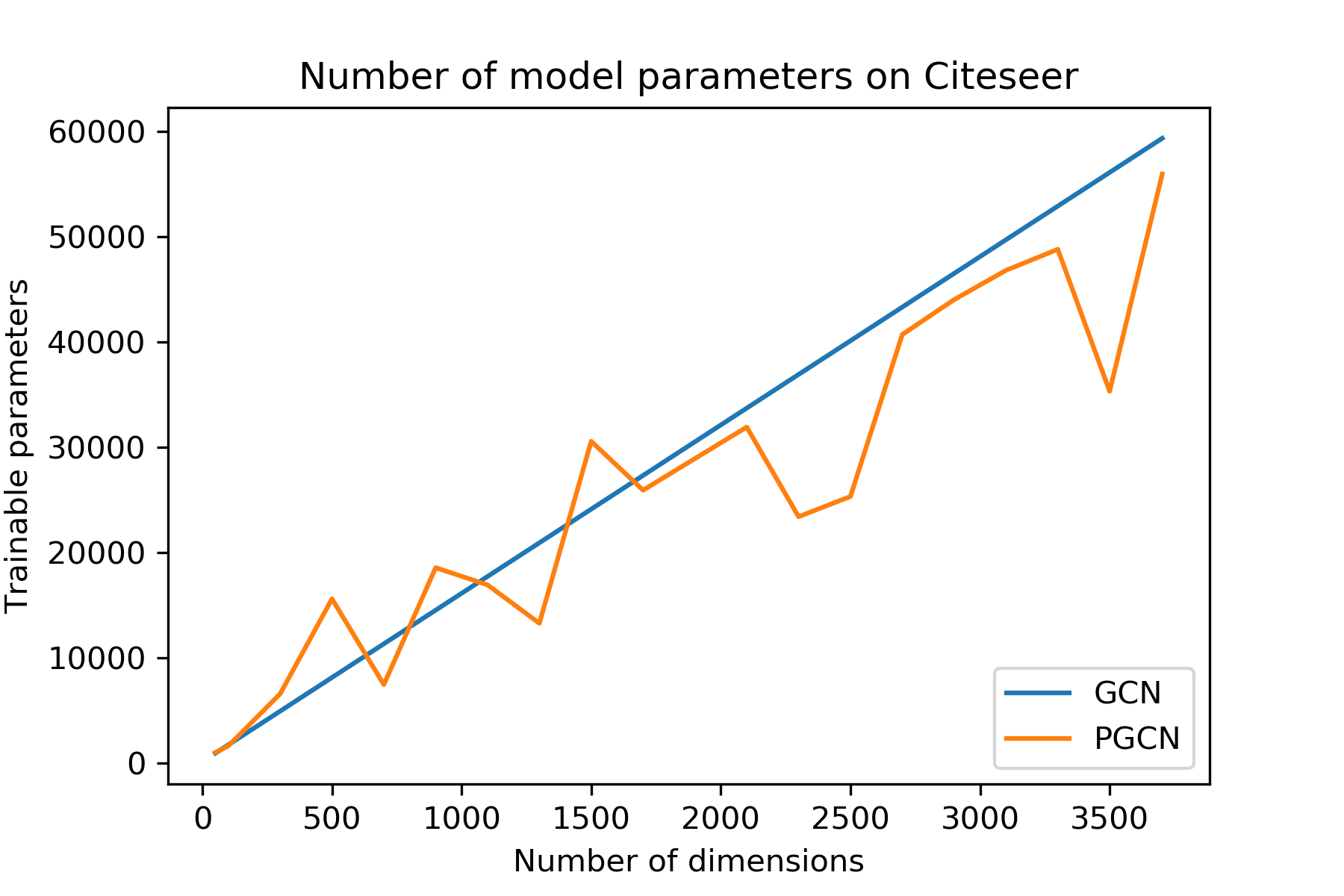}\hfil
        \includegraphics[width=0.33\linewidth]{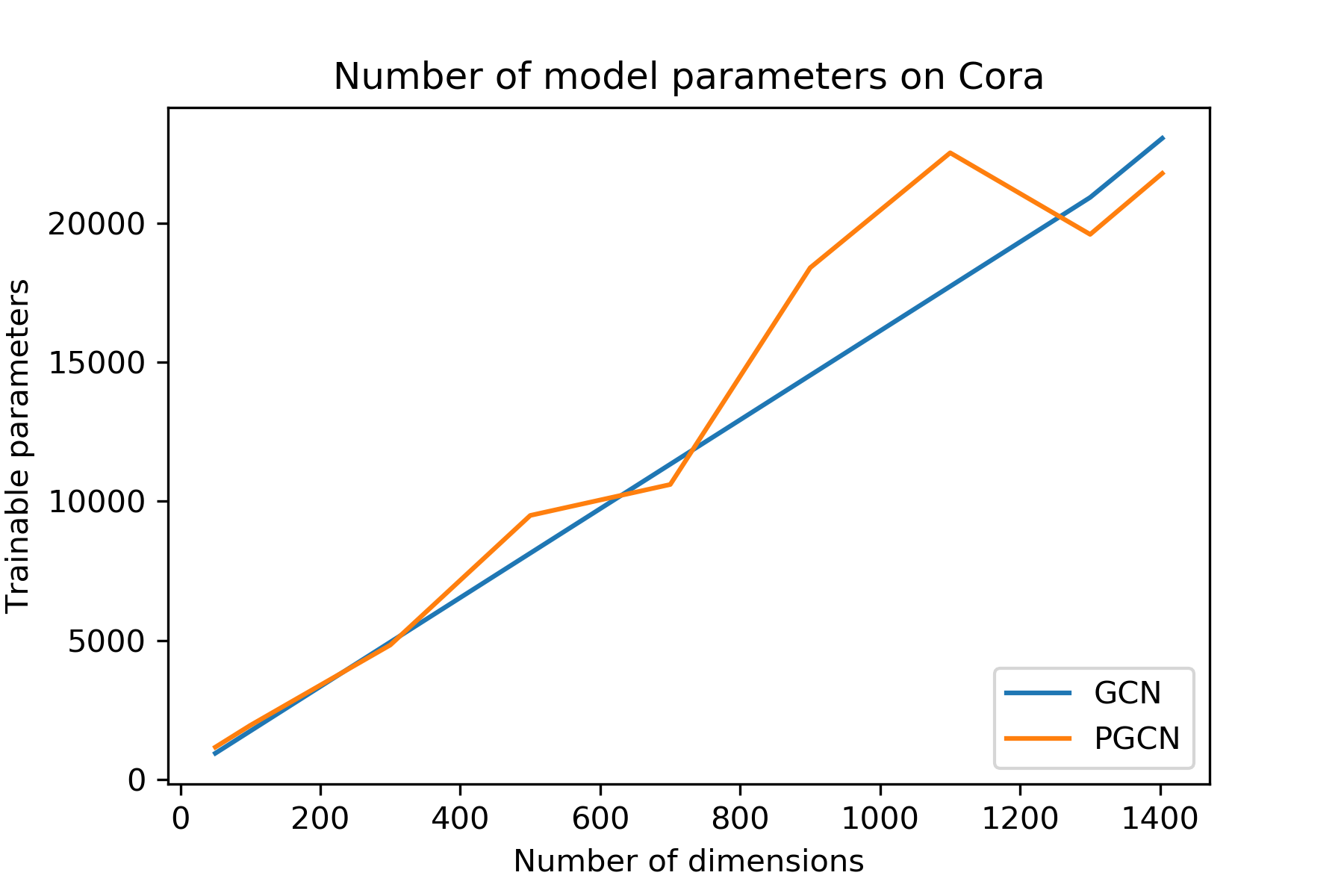}\hfil 
        \includegraphics[width=0.33\linewidth]{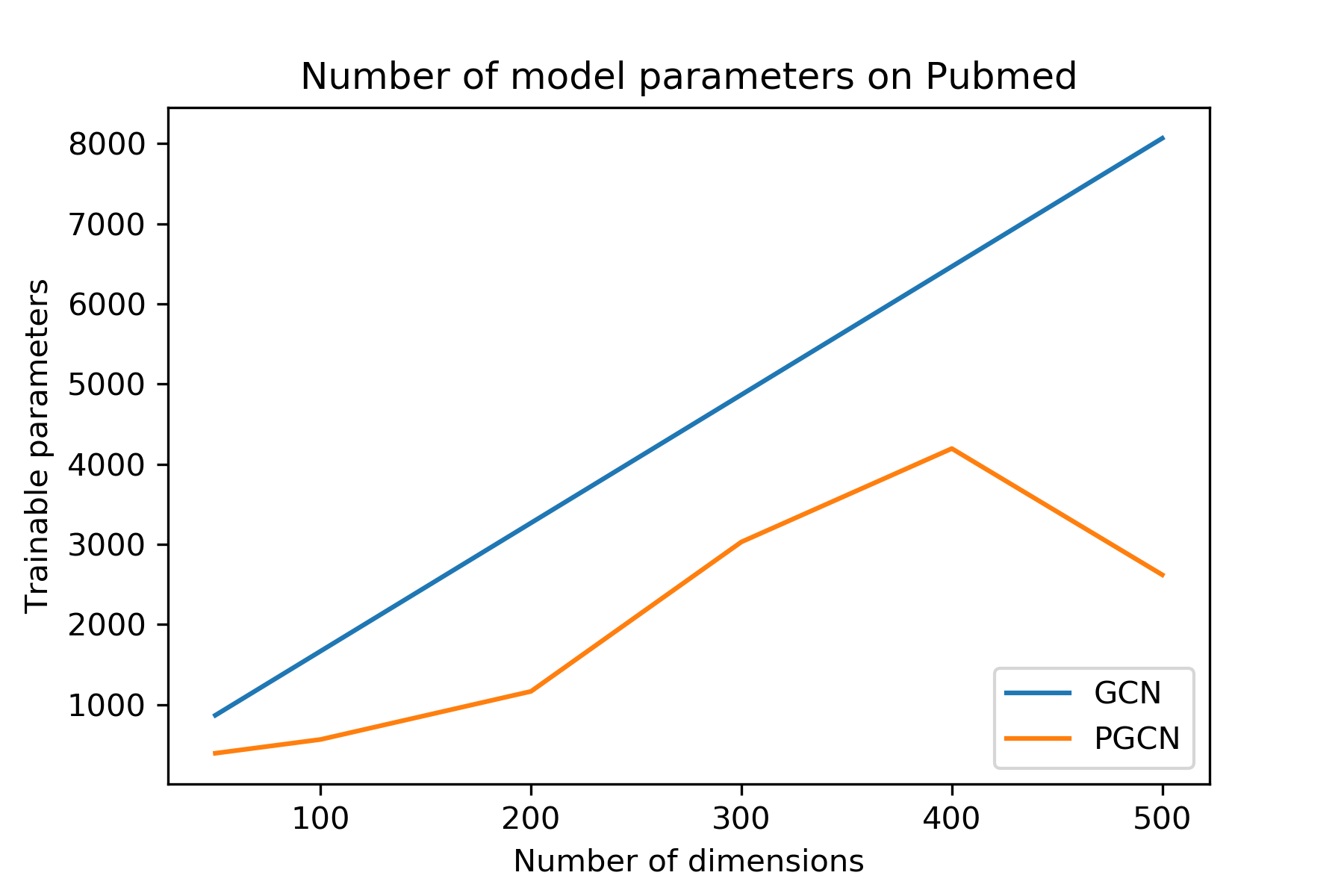}\hfil 
        \caption{Model complexity comparision between GCN and PGCN in different dimensions}
    \end{subfigure}
\caption{Performance and model complexity comparison on citation networks with varying number of dimensions $D^{\prime}$. 
}
\label{fig:randomfeat}
\end{figure*}
Fig. \ref{fig:randomfeat} compares the classification performance of GCN and PGCN methods on all datasets when problems of different difficulty are considered. It can be observed that for all data dimensionalities, our method performs better than the baseline method, while they have a larger difference in classification accuracy in lower dimensionalities, i.e. when the ratio $N_{L}/D$ is higher. This is in line with the findings of \cite{vignac2020choice} indicating that in high-dimensional feature spaces neural network structures tend to perform in similar manners irrespectively to their complexity. On the other hand, when the classification problem is encoded in lower-dimensional feature spaces and, thus, becomes more complex, the structure of the neural network's topology is important. Indeed, as can be observed in Fig. \ref{fig:randomfeat}b, PGCN outperforms GCN with a high margin when the dimensionality $D^{\prime}$ of the node representations is low. The comparison of model complexity with respect to number of trainable parameters in Fig. \ref{fig:randomfeat}c shows that both methods have similar complexity when lower-dimensional feature spaces are used on Citeseer and Cora datasets, while PGCN outperforms GCN in terms of classification perfromance. For Pubmed dataset, PGCN outperforms GCN in terms of both model complexity and classification performance in all cases. 

\section{Conclusion}\label{sec:conclusion}
In this paper, we proposed a method for progressively training graph convolutional networks for semi-supervised node classification which jointly defines a problem-dependant compact topology and optimizes its parameters. The proposed method employs a learning process which utilizes the input to each layer data to grow the network's structure both in terms of depth and width. This is achieved by operating an efficient layer-wise propagation rule leading to a self-organized network structure exploiting data relationships expressed by their vector representations and the adjacency matrix of the corresponding graph structure. Experimental results on three commonly used datasets for evaluating graph convolutional networks on semi-supervised classification indicate that the proposed method outperforms the baseline method GCN and performs on par, or even better, compared to more complex recently proposed methods in terms of classification performance and efficiency.

\appendices

\appendix
\section{Proof of Convergence}\label{S:Appendix}
Here we show that the progressive learning in each layer of the PGCN method converges. Lets assume that $\mathbf{H}_{k}^{(l)} = [\mathbf{h}_{1}^{(l)}, ..., \mathbf{h}_{k}^{(l)}] \in \mathbb{R}^{N\times (D_{1}^{(l)}+...+D_{k}^{(l)})}$ denotes the hidden representations of data produced by using the first $k$ blocks in $l^{th}$ GCN layer and $\mathbf{O}_{k}^{(l)} \in \mathbb{R}^{(D_{1}^{(l)}+...+D_{k}^{(l)}) \times C}$ denotes the finetuned weights connecting all the $k$ blocks of $l^{th}$ layer to the output layer. We prove that the sequence of graph-regularized MSE, $E_{k}^{(l)} \in \mathbb{R}$ obtained with $\mathbf{H}_{k}^{(l)}$ and $\mathbf{O}_{k}^{(l)}$ is monotonically decreasing while it is bounded below by $0$.

Given the fixed hidden representation $\mathbf{H}_{k}^{(l)}$, the finetuned output weights $\mathbf{O}_{k}^{(l)}$ are not necessarily the optimized weights in terms of MSE. 
It can be explained by the following relation: 

\begin{align}
{E}_{k}^{(l)} &= \frac{1}{2} Tr\left(\mathbf{O}_{k}^{{(l)}^{T}} \mathbf{O}_{k}^{(l)}\right) 
+ \frac{\lambda_{1}}{2}\left \| \mathbf{H}_{{L}_{k}}^{(l)} \mathbf{O}_{k}^{(l)} - \mathbf{T}_{L} \right \|_{2}^{2} \nonumber \\
&\geq \frac{1}{2} Tr\left(\mathbf{O}_{k}^{*{(l)}^{T}} \mathbf{O}_{k}^{*(l)}\right) + \frac{\lambda_{1}}{2}\left \| \mathbf{H}_{{L}_{k}}^{(l)} \mathbf{O}_{k}^{*(l)} - \mathbf{T}_{L} \right \|_{2}^{2},
\end{align}
where $\mathbf{O}_{k}^{*(l)}$ denotes the optimized output weights which are obtained by solving the semi-supervised linear regression problem as follows: 
\begin{equation}
\begin{split}
    \mathbf{O}_{k}^{*(l)} &= \left ( (\mathbf{H}_{k}^{(l)})^T \left(\mathbf{I}+\frac{\lambda _{2}}{\lambda _{1}N^{2}}\tilde{\mathbf{L}}\right)\mathbf{H}_{k}^{(l)}+\frac{1}{\lambda _{1}}\mathbf{I}_{{D}^{(l)}}\right )^{-1} \\
    &\cdot(\mathbf{H}_{{L}_{k}}^{(l)})^T\mathbf{T}_{L}.
    \label{eq:optimized_output_W}
\end{split}
\end{equation}

In the next step, when the $({k+1})^{th}$ block is added to the $l^{th}$ layer, the new hidden representation of $l^{th}$ layer would be $\mathbf{H}_{k+1}^{(l)} = \left [ \mathbf{H}_{k}^{(l)}, \mathbf{h}_{k+1}^{(l)} \right]$ in which $\mathbf{H}_{k}^{(l)}$ is fixed from previous step and $\mathbf{h}_{k+1}^{(l)}$ is generated by new randomly initialized weights. The new optimal output weights $\mathbf{\hat{O}}_{k+1}^{(l)} \in \mathbb{R}^{(D_{1}^{(l)}+...+D_{k+1}^{(l)}) \times C}$ which connect the $l^{th}$ layer to output layer is initialized according to (\ref{eq:optimized_output_W}) by substituting $\mathbf{H}_{k}^{(l)}$ by $\mathbf{H}_{k+1}^{(l)}$. The MSE after adding the $(k+1)^{th}$ block would be as follows: 
\begin{align}
\mathcal{E}_{k+1}^{(l)} &=
\frac{1}{2} Tr\left(\mathbf{\hat{O}}_{k+1}^{{(l)}^{T}} \mathbf{\hat{O}}_{k+1}^{(l)}\right) + \frac{\lambda_{1}}{2}\left \| \mathbf{H}_{{L}_{k+1}}^{(l)} \mathbf{\hat{O}}_{k+1}^{(l)} - \mathbf{T}_{L} \right \|_{2}^{2} \nonumber \\
&\leq \frac{1}{2} Tr\left(\mathbf{\tilde{O}}^{T} \mathbf{\tilde{O}}\right) + \frac{\lambda_{1}}{2}\left \| \mathbf{H}_{{L}_{k+1}}^{(l)} \mathbf{\tilde{O}}- \mathbf{T}_{L} \right \|_{2}^{2} \nonumber \\
& \forall \mathbf{\tilde{O}} \in \mathbb{R}^{(D_{1}^{(l)}+...+D_{k+1}^{(l)}) \times C}.
\label{eq:MSE_leastsq}
\end{align}

Since (\ref{eq:MSE_leastsq}) holds for all $\mathbf{\tilde{O}}$, we can replace $\mathbf{H}_{k+1}^{(l)}$, $\mathbf{\tilde{O}}$ with $\left [ \mathbf{H}_{k}^{(l)}, \mathbf{h}_{k+1}^{(l)} \right ]$, $           \begin{bmatrix}
\mathbf{O}_{k}^{*(l)}\\ 
        0
\end{bmatrix}$ respectively to obtain the following relation:  
\begin{align}
\mathcal{E}_{k+1}^{(l)} 
&\leq  \frac{1}{2} Tr\left(
        \left [ \mathbf{O}_{k}^{*(l)^{T}}, 0 \right ]
        \begin{bmatrix}
        \mathbf{O}_{k}^{*(l)}\\ 
        0
        \end{bmatrix} 
        \right) \nonumber \\
        &+ \frac{\lambda_{1}}{2}\left \|
        \left [ \mathbf{H}_{k}^{(l)}, \mathbf{h}_{k+1}^{(l)} \right ]
        \begin{bmatrix}
        \mathbf{O}_{k}^{*(l)}\\ 
        0
        \end{bmatrix} 
 - \mathbf{T}_{L} \right \|_{2}^{2} \nonumber \\
        &= \frac{1}{2} Tr\left(
\mathbf{O}_{k}^{*(l)^{T}}
\mathbf{O}_{k}^{*(l)}
\right) + \frac{\lambda_{1}}{2}\left \|
\mathbf{H}_{k}^{(l)} \mathbf{O}_{k}^{*(l)} 
- \mathbf{T}_{L} \right \|_{2}^{2} \nonumber \\
&\leq {E}_{k}^{(l)}.
\end{align}

After finetuning the network parameters, the output weights are denoted by $\mathbf{O}_{k+1}^{(l)}$ and the MSE would be $E_{k+1}^{(l)}$. It has been proven that stochastic gradient descent converges to a local optimum \cite{robbins1985stochastic} with small enough learning rate, so the following relation holds for the MSE:  

\begin{equation}
    \begin{aligned}
    {E}_{k+1}^{(l)} \leq \mathcal{E}_{k+1}^{(l)}.
    \end{aligned}
    \label{eq:MSE_sgd}
\end{equation}
According to (\ref{eq:MSE_sgd}), (\ref{eq:MSE_conv}) we have the following relation:
\begin{equation}
    \begin{aligned}
    {E}_{k+1}^{(l)} \leq {E}_{k}^{(l)},
    \end{aligned}
    \label{eq:MSE_conv}
\end{equation}
which indicates that the sequence $(E_{k}^{(l)})_{\{k\}}$ is monotonically decreasing. 

Based on the connection of the linear activation function combined with the mean-square error to the soft-max activation function combined with the cross-entropy criterion and maximum likelihood optimization \cite{bishop2007pattern}, an analysis following the same steps as above can be used to show that (when the latter is employed) the sequence $E_{k}^{(l)}$ is also monotonically decreasing. 

\bibliographystyle{IEEEtran}
\bibliography{mybibfile}

\end{document}